# Spatially Constrained Location Prior for Scene Parsing

Ligang Zhang, Brijesh Verma, David Stockwell, Sujan Chowdhury
Centre for Intelligent Systems
School of Engineering and Technology, Central Queensland University
Brisbane, Queensland, Australia
{l.zhang, b.verma, d.stockwell, s.chowdhury2}@cqu.edu.au

*Abstract*—Semantic context is an important and useful cue for scene parsing in complicated natural images with a substantial amount of variations in objects and the environment. This paper proposes Spatially Constrained Location Prior (SCLP) for effective modelling of global and local semantic context in the scene in terms of inter-class spatial relationships. Unlike existing studies focusing on either relative or absolute location prior of objects, the SCLP effectively incorporates both relative and absolute location priors by calculating object co-occurrence frequencies in spatially constrained image blocks. The SCLP is general and can be used in conjunction with various visual feature-based prediction models, such as Artificial Neural Networks and Support Vector Machine (SVM), to enforce spatial contextual constraints on class labels. Using SVM classifiers and a linear regression model, we demonstrate that the incorporation of SCLP achieves superior performance compared to the state-of-the-art methods on the Stanford background and SIFT Flow datasets.

*Keywords—neural networks; image understanding; scene parsing; image segmentation; object classification*

## I. Introduction

Scene or image parsing is a process of classifying every pixel of a query image into a correct sematic category, such as tree, grass and sky. It is a critical task in computer vision and has a wide range of potential applications, such as automatic vehicle navigation and hazard object identification. The main challenge of scene parsing is to generate effective modelling of the visual variability of a considerable amount of both structured and unstructured object classes while being robust against impacting factors such as viewpoint, illumination, size and occlusion [1].

The problem of scene parsing has to address several processing tasks, including the choice of elementary regions (e.g. pixel, patch and superpixel), the selection of visual features to characterize them (e.g. color, shape and location), the design of supervised or unsupervised classifiers for obtaining semantic confidence, the extraction of contextual features, and the integration of prediction models in considering contextual information. Recent advances in scene parsing have advocated the use of superpixels, which contain a set of pixels sharing similar appearance characteristics. A parametric or non-parametric prediction procedure is often incorporated for class label assignment based on a set of superpixel-level visual features. However, visual feature based prediction shows only limited capability to effectively handle the complexity of objects in natural scene parsing, as it treats each individual superpixel independently without taking into account the scene context, such as inter-class spatial relationships. For instance, cars and roads have a high probability of co-existing in a highway scene and cars are often above roads, but cars are unlikely to be present in a sea scene.

To reduce the amount of ambiguity and misclassification in the results of visual feature based prediction, increased attention has been shifted towards modelling and incorporating the scene context at two stages: 1) class label inference which often adopts graphical models to impose contextual consistency of class categories in the scene. The widely used graphical models include Conditional Random Fields (CRFs) [2], [3], [4], Markov Random Fields (MRFs) [5], and region boundary based energy function [6]; and 2) contextual feature extraction which tries to capture the intrinsic correlations between objects embedded in different types of scenes. The features include absolute location [1], relative location [7], directional spatial relationships [8], object co-occurrence statistics [9], etc. These graphical models and contextual features have been shown to greatly improve the performance of approaches using visual features alone. However, the modelling of contextual information in complicated natural scenes is still far from perfect, particularly for large datasets with varying properties. How to design useful global and local contextual features and how to effectively incorporate them in enforcing constraints on class labelling are still largely unresolved challenges.

In this paper, we propose a new contextual feature - Spatially Constrained Location Prior (SCLP) for scene parsing in complex natural images. The SCLP is inspired by the recent success of using relative or absolute location priors of objects for effective image understanding. Unlike previous works [1], [7], the SCLP takes into account both relative and absolute location priors, and achieves a good trade-off between them by collecting object co-occurrence frequencies in spatially constrained image blocks. In addition, it also partially encodes the directional spatial relationships and thus contains rich contextual cues useful for enforcing constraints on labels of complicated objects. For a testing superpixel from a spatial block, its global and local contextual information is obtained by performing class label preference votes from superpixels in other blocks and adjacent superpixels respectively. To demonstrate the effectiveness of the SCLP, we integrate it in a linear regression model to refine the class labels from visual feature based prediction, showing superior performance on two publicly available benchmark datasets – Stanford background and SIFT Flow.

The rest of the paper is organized as follows. Section II presents related work. Section III describes the proposed SCLP approach. Section IV presents the experiments and results. Finally the conclusions and future research are presented in Section V.

## II. RELATED WORK

Scene parsing has been investigated extensively in previous studies. Early approaches obtain the class label for each individual pixel using a set of low-level visual features extracted at each pixel [1], [6] or a local patch around each pixel [10]. A supervised machine learning process is often employed to establish the correlation between visual features and class labels. With the increased complexity of natural scenes, such low-level analysis techniques often face a big challenge for high-level image understanding, as pixel-level features cannot capture more robust statistics about the appearance of a local region, patch-level features are prone to noise from background objects, and both features are unable to consider the global context.

Recent advances [4], [11], [12], [13] on scene parsing have shifted to the adoption of superpixel-level visual features and the integration of semantic and structural contextual information to improve the classification results. Oversegmented superpixels, which group pixels into perceptually meaningful atomic regions, have advantages of coherent support regions for a single labelling on a naturally adaptive domain rather than on a fixed window, more consistent feature extraction capturing contextual neighbouring information by pooling over feature responses from multiple pixels, and significant reduction of complexity and computation. The most widely used superpixel-level features include color [1], [7], [11], [14], [15], [16], texture [1], [7], [9], [12], [14], [15], boosted classifier scores, appearance [14], shape, location, etc. It is worth mentioning that two typical sets of superpixel-level visual features are available in the STAIR Vision Library [11] and SuperParsing [14].

Contextual information has shown a crucially important role in refining semantic labelling accuracy and boosting scene parsing performance. Studies [17] found that the human classification of superpixels has lower accuracy than machine classification when pixels outside superpixels are invisible, implying the importance of context in assisting human recognition of objects in scenes. One popular method of incorporating context is to use graphical models, such as CRFs [2], [3], [4], MRFs [5] and energy minimization functions [6], which enforce the spatial consistency of category labels between neighbouring superpixels (or pixels). However, one problem of using superpixels is that label purity of superpixels cannot be guaranteed due to the difficulty of perfect image segmentation. On one hand, the size of superpixels should be small to avoid straddling object boundaries. On the other hand, the size should be large enough to provide reliable features. To address the problem, hierarchical models have been proposed to generate a pyramid of image superpixels and perform classification optimization over multi-levels of images to alleviate the effect of inaccurate region boundaries and incorporate features from large-scale regions. These models include hierarchical CRF [18], multi-scale convolutional neural networks [4], recursive context propagation network [13], [19], stacked hierarchical learning [15], pylon model [18] and histograms in superpixel neighbourhoods [12].

Unlike graphical models that enforce contextual constraints directly at the class label inference stage, the context has also been investigated in terms of a rich set of semantic descriptions. The most widely adopted contextual features include absolute location [1], which captures the dependence of class labels on the absolute location of pixels in the image, relative location [7], which represents the relative location offsets between objects in a virtually enlarged image, directional spatial relationships [8] which encode the beside, below, above, enclosed spatial arrangements of objects, object co-occurrence statistics [9],[20] which reflect the likelihood of two objects co-existing in the same scene, as well as global and local context descriptors [21]. A drawback of the relative location, directional spatial relationships and object co-occurrence statistics is that they completely discard absolute spatial coordinates of objects in the image, and thus they cannot capture spatial contextual information such as sky often appears in the top part of a scene. By contrast, the absolute location excessively retains all pixel coordinates of objects, and thus it requires a large amount of training data to collect reliable prior statistics for each object and for each image pixel. The proposed SCLP is aimed at overcoming the drawbacks of these contextual features and enforcing more robust constraints on label consistency of superpixels by introducing spatially constrained blocks.

The work that is most similar to ours is [7], which integrated the relative location offsets between each pair of objects in a probabilistic classification model to refine the results of superpixel-level visual feature based prediction. Like our approach, they generated spatial relationships between classes from the training data and utilized them as contextual information for multi-class scene parsing. Unlike our approach, the spatial relationships in their method were based on relative location offsets of objects, which completely discard absolute coordinates of objects that also convey useful contextual information for scene parsing. Our approach considers both relative and absolute locations and achieves a trade-off between them by collecting object co-occurrence frequencies in spatially constrained image blocks.

## III. PROPOSED APPROACH

As shown in Fig. 1, the proposed approach is composed of four main processing steps: 1) generation of Spatially Constrained Location Prior (SCLP), 2) prediction of class probabilities using visual feature based classifiers, 3) propagation of contextual class votes based on SCLP, and 4) integration of visual feature based class probabilities and contextual class votes.

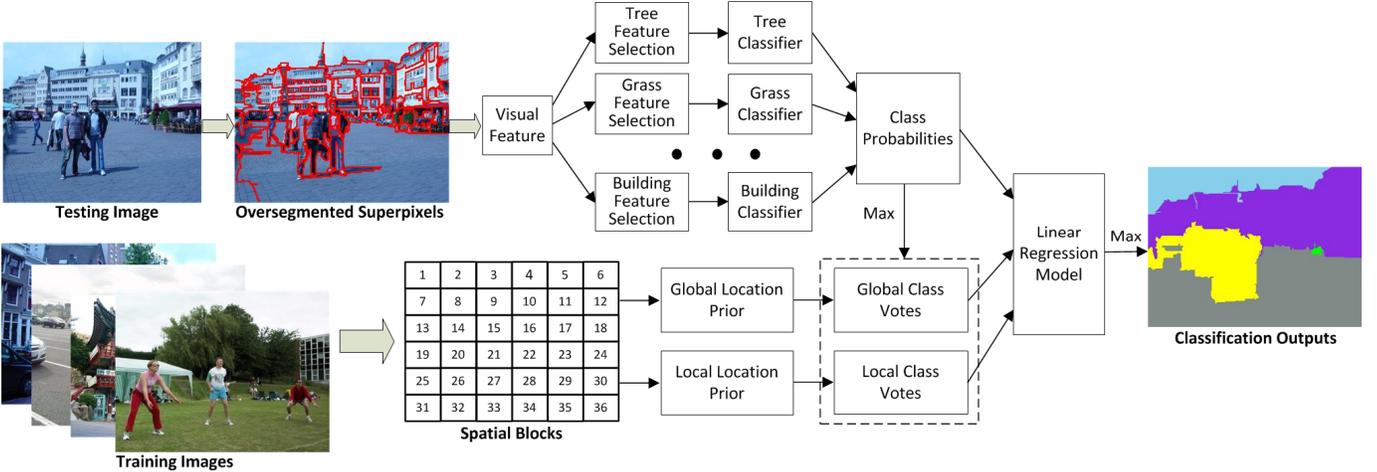

Fig. 1. Framework of the proposed approach for scene parsing (best reviewed in color). The SCLP features are learnt from all training images by collecting global and local location priors of objects in equally divided blocks. For each superpixel testing, we obtain: 1) class probabilities predicted using visual feature based classifiers, 2) global class votes obtained by casting votes from superpixels in all other spatial blocks based on global location prior, and 3) local class votes obtained by casting votes from adjacent superpixels based on local location prior. The class probabilities, global and local class votes are integrated using a linear regression model to approximate their correlation, and a majority voting strategy is used to assign a class label to each superpixel.

The approach first segments all images into a set of homogeneous superpixels using a fast graph-based algorithm [22] and takes superpixels as basic processing units. At the training stage, the SCLP is formed by dividing training images into equally sized spatial blocks and then calculating both global and local object co-occurrence frequencies in all blocks, which encode the likelihood of classes concurrently appearing in certain spatial parts of the images and reflect prior inter-class spatial relationships inherently embedded in different types of training scenes. For a testing image, we first predict the probabilities of all classes for each superpixel independently using visual feature based class-specific classifiers. The most probable class for each superpixel is then obtained and further combined with the SCLP to learn image-dependent contextual class votes, which reflect class label preferences for each superpixel voted by the global and local context respectively in the testing image. Visual feature based class probabilities and contextual class votes are integrated using a linear regression model to approximate their best relationships, yielding a probability for each class. A class label is finally decided for each superpixel using a majority voting on the probabilities of all classes.

### A. Spatially Constrained Location Prior (SCLP)

The SCLP encodes the inter-class spatial correlations with regard to their prior class label distributions in spatially constrained blocks of images. Four types of inter-class spatial correlations are considered in the SCLP, including object co-occurrence statistics, relative location (e.g. distance between objects) and absolute location (e.g. top or middle part of the scene), and directional spatial relationships (left or right side of an object). These spatial correlations often commonly exist in natural scenes and carry important semantic contextual information about the scene, and thus can be used to enforce useful contextual constraints on the class labels predicted using visual feature based classifiers to refine scene parsing results.

The calculation of SCLP is explained as follows. Let $I(v) \in R^3$ be an image defined on a set of pixels $v$, $C = \{c_i | i = 1,2,...,M\}$ indicate the class labels, $S(v) = \{s_j | j = 1,2,...,N\}$ denote the set of superpixels oversegmented from $I$, and $M$ and $N$ are the number of all classes and superpixels respectively. To consider both relative and absolute location offsets of objects, we first partition $I$ into a set of equally distributed blocks $\{I = \bigcup(B_k), k = 1,2,...,K\}$ and $K$ is the number of blocks. The use of blocks is aimed at keeping a good trade-off between relative and absolute location in the sense that the relative location of objects is encoded by spatial relationships between blocks, while the absolute location is preserved by spatial coordinates of blocks. The spatial distributions of all blocks also preserve the directional spatial relationships between objects within different blocks. As objects can appear at any location in a scene, both global and local location priors are created to represent the likelihood of two objects co-occurring in either a long or a short distance apart in the scene, and they capture the global and local context respectively.

(1) Global location prior. Assume a pixel with a class label $\acute{c}$ occur in a block $B_{k1}$, a matrix $\mathcal{M}_{c|\acute{c}}(k_1, k_2)$, $k_1 \neq k_2$ indicates the probability that a pixel with a class $c$ occurs in a block $B_{k2}$ and $\mathcal{M} \in \mathcal{R}^{M*M*K*(K-1)}$. The matrix is normalized to ensure a conditional probability distribution over all classes in each block, i.e. $\sum_{c=1}^{M} \mathcal{M}_{c|\acute{c}}(k_1, k_2) = 1$. For each class and each block, a matrix with $M*(K-1)$ elements can be generated. The matrix is learnt for each block independently from the training data. For a superpixel in a block $B_{k1}$, the global location prior reflects the confidence in its class label with the support of contextual information from all superpixels in other $K-1$ blocks $\bigcup(B_{k2}) k_2 \neq k_1$.

(2) Local location prior. Given a superpixel $s_j$ with a class label $\acute{c}$, a matrix $\bar{\mathcal{M}}_{c|\acute{c}}(s_j, s_p)$ represents the probability that one of its adjacent superpixel $s_p \in \bar{S}_j$ with a class $c$ and $\bar{\mathcal{M}}_{c|\acute{c}}(s_j, s_p) \in \mathcal{R}^{M*M}$. The local matrix is calculated for each

pairs of adjacent superpixels in the training data and the spatial block is not considered during the calculation so that all superpixel pairs can appear at any location of the scene. It compensates for the "self-support" contextual information for superpixels within each block that is not considered in the global location prior. For a superpixel, the local location prior accounts for the confidence in its class label with the support of contextual information from its neighbouring superpixels.

The occurrence frequency of each class is calculated by counting pixels within all superpixels labelled to this class. The class label of each superpixel is set to be the majority vote of the ground truth pixel labels. Due to variations in the shape and size of segmented superpixels, pixels from a superpixel may be distributed across the boundaries of multiple spatial blocks. Thus we assign each superpixel in an image to a unique block based on the centroid of all pixels within the superpixel.

### B. Visual Feature Based Class Probabilities

For a testing image, we extract a set of visual features from all superpixels and then employ a supervised learning process, which generates multiple class-specific classifiers to obtain an approximate prediction of the probabilities of all superpixels belonging to each class. The most probable class is chosen for each of all superpixels by taking the maximum probability and further used to generate contextual class votes (in next subsection).

(1) Superpixel-level visual features. Extracting a discriminative and representative set of visual features is a central step in obtaining accurate predictions of class probabilities of superpixels. Following [14], we extract a set of superpixel-level color, geometric and texture features. The features include the mean and standard deviations of RGB colors over the pixels of each superpixel (2*3 dimensions), the top height of superpixel bounding box to the image height (1 dimension), the mask of superpixel shape over the image (8*8 dimensions), 11-bin histograms of RGB colors (11*3 dimensions), 100-bin histogram of textons (100 dimensions), and 100-bin histogram of dense SIFT descriptors (100 dimensions) over the superpixel region. In addition, we also obtain RGB, texton and SIFT histograms (233 dimensions) over the superpixel region dilated by 10 pixels. The final visual features are composed of 6+1+64+33+200+233=537 elements.

(2) Class-specific feature selection. Given a pre-partitioned image with superpixels $S(v) = \{s_j | j = 1,2, ..., N\}$ and the corresponding visual features $F^v = \{f_j^v | j = 1,2, ... N\}$, we employ a feature selection process to obtain class-specific feature subsets $\hat{X}^l = \{\hat{f}_{i,j}^v | i = 1,2, ... M; j = 1,2, ... N\}$ for each of all classes using the minimum redundancy maximum relevance (mRMR) algorithm [23]. In natural scenes, objects may be defined by different attributes and thus best represented by class-specific features to differentiate one class from others.

(3) Class probability prediction. We incorporate a supervised train-test procedure to accomplish the multi-class probability prediction task. Instead of training a single multi-class classifier for all classes, we choose to train a series of one-vs-all binary classifiers for each class. For the $j^{th}$ superpixel $s_j$, we obtain its class probability for the $i^{th}$ class $c_i$:

$$P^v(c_i|s_j) = \emptyset_i(\hat{f}_{i,j}^v) \quad (1)$$

where, $\hat{f}_{i,j}^v$ is the visual feature subset of $s_j$ for the $i^{th}$ class $c_i$, and $\emptyset_i$ is the trained binary classifier for $c_i$. In principle, the classifier can be any type of machine learning algorithms with probabilistic prediction capabilities, such as Artificial Neural Networks (ANNs) and Support Vector Machine (SVM). For all $M$ classes, we can get a class probability vector for $s_j$:

$$P^v(C|s_j) = [P^v(c_1|s_j), ..., P^v(c_i|s_j), ..., P^v(c_M|s_j)] \quad (2)$$

The above vector contains the likelihood of each superpixel $s_j$ belonging to all classes $C$, and then we assign $s_j$ to the class which has the maximum probability:

$$s_j \in \hat{c} \quad if \quad P^v(\hat{c}|s_j) = \max_{1<i<M}(P^v(c_i|s_j)) \quad (3)$$

The use of class-specific classifiers brings three advantages: 1) supporting the selection of class-specific features, which allows the use of most discriminative features specifically for each class, 2) focusing on training a classifier for a specific class at each time, which is often more effective than training a single multi-class classifier, particularly for datasets with a large number of classes, and 3) handling the problem of unbalanced training data between classes, particularly for natural data where there are many rarely occurring but important classes. In real-world scene datasets, the distribution of pixels is likely to be heavy-tailed to several common classes and there are only a very limited number of pixels for rare classes. In such cases, using a multi-class classifier has the risk of completely ignoring rare classes and being favourably biased towards common classes.

### C. Contextual Class Votes Based on SCLP

Based on visual feature based class probabilities obtained for each superpixel, this part describes how to incorporate the global and local location priors into assisting the prediction of class probabilities for superpixels in a testing image. We first obtain the most probable class for each of all superpixels and then adopt a voting strategy to obtain contextual class votes for each superpixel based on the global and local location priors.

Given a superpixel $s_j$ in a block $B_{k2}$, its most probable class $\hat{c}$ and class probability $P^v(\hat{c}|s_j)$, it receives Q votes from all superpixels $S = \{s_q | q = 1,2, ..., Q\}$ in all the rest of blocks:

$$V^g(C|s_j) = \sum_{k_1 \neq k_2; s_q \in \hat{c}} w_q \times \mathcal{M}_{c|\hat{c}}(k_1, k_2) \quad (4)$$

where, $w_q = P^v(\hat{c}|s_j) * \mathbb{C}(s_q)$ is a weight given to the vote from the superpixel $s_q$ and $\mathbb{C}(s_q)$ is the total number of pixels in $s_q$. Because $s_q$ can have $M$ possible class labels, the resulting $V^g(C|s_j)$ can be written as:

$$V^g(C|s_j) = [V^g(c_1|s_j), V^g(c_2|s_j), ..., V^g(c_M|s_j)] \quad (5)$$

where, $V^g(c_i|s_j)$ is the contextual label votes of being the $c_i$ class for $s_j$ based on the global location prior.

Let $\bar{S}_j = [s_p] \, p = 1,2, ..., P$ be the set of adjacent superpixels of $s_j$ and $s_p$ be the $p^{th}$ member of $\bar{S}_j$, the $s_j$ receives $P$ votes from all its neighbours $s_p$:

$$V^l(C|s_j) = \sum_{1<p<P; s_p \in \hat{c}} w_p \times \bar{\mathcal{M}}_{c|\hat{c}}(s_j, s_p) \quad (6)$$

where, the weight $w_p$ is calculated the same way as $w_q$ in (4). In a similar way as in (5), the contextual label votes of all classes for $s_j$ can be obtained based on the local location prior:

$$V^l(C|s_j) = [V^l(c_1|s_j), V^l(c_2|s_j), ..., V^l(c_M|s_j)] \quad (7)$$

Now, we have calculated two contextual class votes for a testing superpixel based on the global and local location priors respectively, and they reflect the confidence in all classes for the superpixel using global and local context respectively in the testing scene.

### D. Probabilistic Object Classification

The probabilistic object classification is the last step of the proposed approach and it aims to produce context-sensitive classification of each superpixel in a testing image through the incorporation of a unified regression model to seamlessly integrate visual feature based class probabilities $P^v(C|s_j)$, local class votes $V^l(C|s_j)$ and global class votes $V^g(C|s_j)$. Towards this aim, we obtain class-specific optimized coefficients of a linear regression model to best describe the correlations between each class label and the three predictive terms. These coefficients are anticipated to inherently account for different contributions of visual features and contextual cues in predicting the class labels of superpixels in the testing image.

To keep consistency with class probabilities $P^v(C|s_j)$, both global and local class votes $V^g(C|s_j)$ and $V^l(C|s_j)$ are converted into probabilities $P^g(c_i|s_j)$ and $P^l(c_i|s_j)$ respectively by dividing the sum of their elements. A linear regression model is then employed to integrate all three types of probabilities for the prediction of the probability of the $i^{th}$ class $c_i$ for superpixel $s_j$:

$$P(c_i|s_j) = w_i^c + w_i^v \times P^v(c_i|s_j) + w_i^l \times P^l(c_i|s_j) + w_i^g \times P^g(c_i|s_j) \quad (8)$$

where, $w_i^c$ is a constant value, and $w_i^v$, $w_i^l$ and $w_i^g$ are coefficients for $P^v(c_i|s_j)$, $P^l(c_i|s_j)$ and $P^g(c_i|s_j)$ respectively. The coefficients are learnt to maximize the likelihood score over the labelled training data.

The superpixel $s_j$ is finally assigned to the class $\bar{c}$ which has the maximum probability across all classes using a majority voting strategy:

$$s_j \in \bar{c} \text{ if } P(\bar{c}|s_j) = \max_{1 < i < M} P(c_i|s_j) \quad (9)$$

## IV. EXPERIMENTS

In this section, we evaluate the performance of the proposed approach on two widely used benchmark datasets for scene parsing: Stanford background and SIFT Flow. We use two evaluation metrics: *global accuracy* which is the ratio of correct pixels to the total pixels in all testing images, and *class accuracy* which is the mean of category-wise pixel accuracy. They are dominated by the most common and rarest classes respectively. We also compare our results with the accuracies reported in state-of-the-art scene parsing algorithms.

### A. System Parameter Settings

The parameters of the graph-based image segmentation algorithm are set based on [14], i.e. $\sigma = 0.8$, $min = 100$, $k = 200 * \max(1, sqr(D_I/640))$, and $D_I$ is the larger dimension (height or width) of an image $I$. For visual feature selection using the mRMR algorithm, all continuous values are discretised into three states of -2, 0 and 2, and the length of the selected features is fixed to 50 for all classes. The SVM with a RBF kernel is used as the classifier. A total number of 36 spatial blocks are used, i.e. 6 in width and 6 in height.

### B. Stanford Background Dataset

The Stanford background dataset [11] comprises of 715 images of outdoor scenes assembled from existing public datasets, including LabelMe, MSRC, PASCAL and Geometric Context. There are eight object classes, including sky, tree, road, grass, water, building, mountain and foreground object. All images have approximately 320*240 pixels, with each containing at least one foreground object. All image pixels are manually annotated into one of eight classes or unknown object using Amazon Mechanical Turk. Following the evaluation procedure in [11], five-fold cross validations are conducted to obtain average classification accuracy: 572 images are randomly selected for training and 143 images for testing in each of five folds.

Table I shows the accuracy of the proposed approach compared with reported accuracies by state-of-the-art methods. The proposed approach achieves global accuracies of 81.2%, which is comparable to the state of the art performance. A lower class accuracy of 71.8% is observed for our approach compared with 76.0% and 79.1% accuracies by existing methods [4], [13], indicating that the proposed approach tends to focus on common classes with a large proportion of training pixels, and this is within our expectation as the SCLP was generated based on class pixel distributions on the training data. For a specific class, the more training pixels, the more reliable its SCLP features and the more accurate the corresponding contextual class votes. The proposed approach significantly outperforms using visual features alone with increases of more than 38% and 36% for global and class accuracies respectively, confirming the important role of the SCLP contextual features in reducing misclassification and refining classification results.

Table II presents the confusion matrix for eight classes. It can be seen that sky, building and road are three easiest classes for correct classification with more than 87% accuracies, whereas mountain and water are two most difficult classes with less than 56% accuracies. The results agree with those in previous studies [11], [15], [26], where sky and mountain had the highest and lowest accuracies respectively (approximately 92% and 14%). A significant proportion (24.6%) of mountain pixels are misclassified to tree, probably due to the overlap in color features. Similarly, a considerable amount (20% and 17.3%) of water pixels are misclassified to road and foreground categories. Among all classes, the foreground leads to the most overall misclassification to other classes, and this is largely due to big variations in the appearance of foreground objects in natural scenes.

TABLE I.  PERFORMANCE (%) COMPARISONS WITH STATE-OF-THE-ART METHODS ON THE STANFORD BACKGROUND DATASET

| Ref. | Global Acc. | Class Acc. |
|---|---|---|
| Munoz et al. 2010 [15] | 76.9 | 66.2 |
| Kumar et al. 2010 [24] | 79.4 | - |
| Lempitsky et al. 2011 [18] | 81.9 | 72.4 |
| Farabet et al. 2013 [4] | 81.4 | 76.0 |
| Sharma et al. 2014 [19] | 81.8 | 73.9 |
| Shuai et al. 2015 [25] | 81.2 | 71.3 |
| Sharma et al. 2015 [13] | **82.3** | **79.1** |
| Visual feature | 43.1 | 35.2 |
| Proposed SCLP | **81.2** | **71.8** |

TABLE II.  CONFUSION MATRIX FOR EIGHT OBJECTS ON THE STANFORD BACKGROUND DATASET

|  | Sky | Tree | Road | Grass | Water | Bldg. | Mtn. | Fgnd. |
|---|---|---|---|---|---|---|---|---|
| Sky | **91.2** | 3.9 | 0.1 | 0.1 | 0.1 | 3.3 | 0.1 | 1.2 |
| Tree | 2.7 | **74.3** | 1.1 | 1.2 | 0.2 | 14.4 | 0.2 | 6.0 |
| Road | 0.1 | 0.6 | **87.2** | 1.8 | 0.3 | 2.8 | 0 | 7.1 |
| Grass | 0.3 | 5.2 | 10.6 | **64.4** | 1.0 | 1.8 | 0.8 | 15.9 |
| Water | 3.2 | 0.6 | 20.0 | 1.7 | **55.6** | 1.0 | 0.7 | 17.3 |
| Bldg. | 1.4 | 4.5 | 1.3 | 0.2 | 0.3 | **88.6** | 0.1 | 3.7 |
| Mtn. | 7.6 | 24.6 | 4.7 | 4.7 | 2.0 | 8.7 | **37.2** | 10.5 |
| Fgnd. | 1.3 | 4.2 | 6.1 | 1.1 | 0.7 | 11.1 | 0.1 | **75.4** |

## C. SIFT Flow Dataset

The SIFT Flow dataset [27] includes 2,688 images that are thoroughly labelled by LabelMe users. Most of the images are outdoor scenes and the top 33 object categories with the most labelled pixels are used in the dataset, including sky, building, mountain, tree, road, etc. There is also an additional 'unlabelled' category for pixels that are not labelled or labelled as other object categories. All images have a resolution of 256*256 pixels. In our experiments, the commonly adopted train/test data split introduced in [27] is used: 2,488 training images and 200 test images.

Table III presents accuracy comparisons of the proposed approach with state-of-the-art methods. The proposed approach shows superior performance over existing methods and it significantly improves the state-of-the-art global accuracy from 80.9% to 87.0% and has one of the top class accuracies of 36.9%. One possible reason for the significant improvement is that there is a much larger number of training images on the SIFT Flow dataset (compared to the Stanford background dataset), which is crucial for collecting reliable SCLP contextual features for all classes. The proposed approach also has more than 23% and 20% higher global and class accuracies respectively than using visual feature alone. The results indicate the great advantage of incorporating SCLP features in handling the complexity of objects in natural scene parsing. Fig. 2 visually compares the classification results on sample images between the proposed approach and using visual features alone, from which we can see the significant role of SCLP in removing misclassification.

TABLE III.  PERFORMANCE (%) COMPARISONS WITH STATE-OF-THE-ART METHODS ON THE SIFT FLOW DATASET

| Ref. | Global Acc. | Class Acc. |
|---|---|---|
| Farabet et al. 2013 [4] | 78.5 | 29.6 |
| Tighe & Lazebnik 2013 [28] | 78.6 | 39.2 |
| Singh & Kosecka 2014 [29] | 79.2 | 33.8 |
| Yang et al. 2014 [21] | 79.8 | **48.7** |
| Sharma et al. 2014 [19] | 79.6 | 33.6 |
| Najafi et al. 2015 [30] | 76.6 | 35.0 |
| Nguyen et al. 2015 [31] | 78.9 | 34.0 |
| Sharma et al. 2015 [13] | **80.9** | 39.1 |
| Visual feature | 63.4 | 16.2 |
| Proposed SCLP | **87.0** | **36.7** |

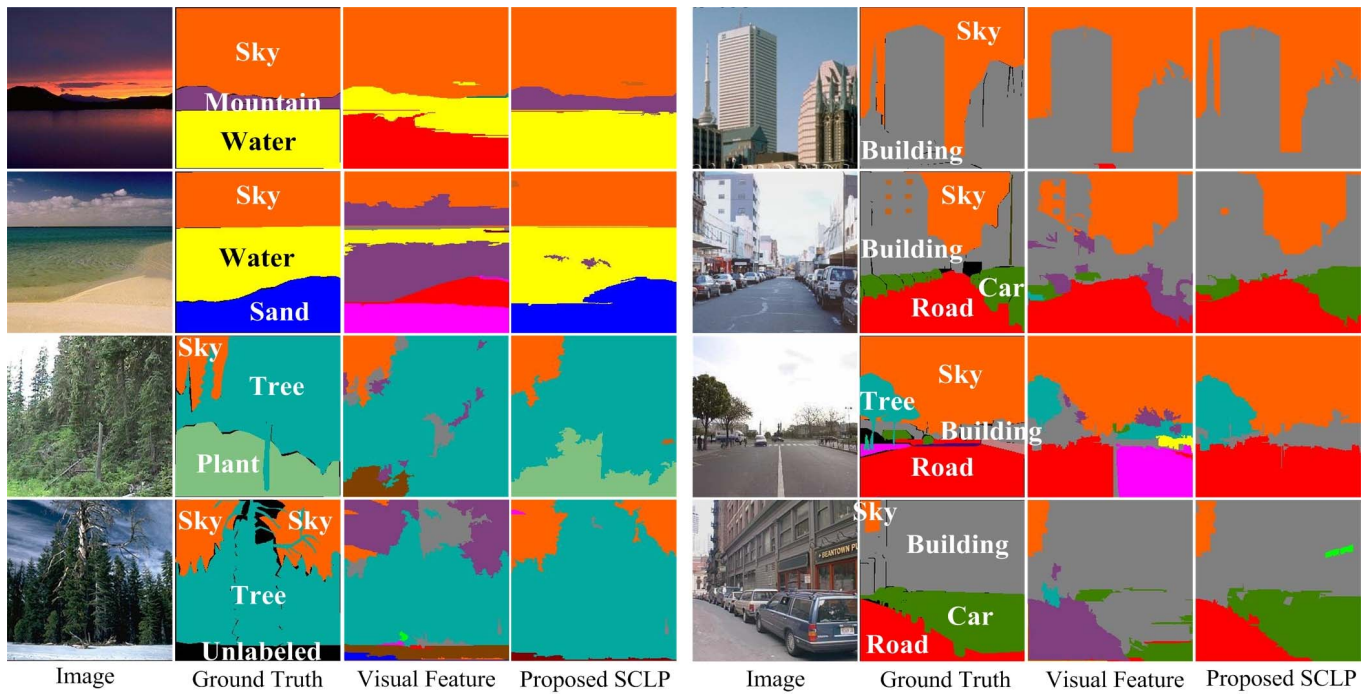

Fig. 2. Qualitative results on the SIFT Flow dataset (best reviewed in color). Compared with using visual features alone, the proposed approach shows more robust scene parsing results due to the adoption of contextual SCLP features, with many misclassification successfully removed.

## V. CONCLUSIONS

This paper presents a novel and effective contextual feature - spatially constrained location prior for scene parsing in complex natural images. The SCLP feature incorporates spatially constrained blocks to collect robust class co-occurrence statistics from the training data by pooling over pixels in each block, while achieving a good trade-off between absolute and relative location prior of objects via retaining spatial correlations between blocks. It has the advantage of capturing a richer set of both global and local contextual information embedded in each type of scene, and thus is very useful to remove misclassification in the results of visual feature based prediction. It is general and can be combined with various visual feature based probabilistic prediction models, such as artificial neural networks and support vector machine, for robust scene parsing. We demonstrate that the incorporation of the SCLP feature brings significant improvements to accuracy over using visual features alone, and achieves the recorded accuracy of 87.0% on the SIFT Flow dataset and one of the top accuracies of 81.2% on the Stanford background dataset. To further improve the accuracy, our future work will investigate the impact of different parameters on the performance, such as the size of spatial blocks, the number of superpixels, the dimension of selected features, as well as exploit other techniques for integrating visual and contextual features, such as logistic regression models.

## ACKNOWLEDGMENT

This research was supported under Australian Research Council's Linkage and Discovery Projects funding scheme (project numbers LP140100939 and DP160102639).